\documentclass{article}

\usepackage{microtype}
\usepackage{graphicx}
\usepackage{subfigure}
\usepackage{booktabs} 

\usepackage{hyperref}
\usepackage{fancyhdr}

\usepackage[accepted]{icml2023}

\usepackage{amsmath}
\usepackage{amssymb}
\usepackage{mathtools}
\usepackage{amsthm}
\usepackage{multirow}

\usepackage[capitalize,noabbrev]{cleveref}

\theoremstyle{plain}

\theoremstyle{definition}

\theoremstyle{remark}

\usepackage[textsize=tiny]{todonotes}

\begin{document}

\twocolumn[
\icmltitle{Slot-VAE: Object-Centric Scene Generation with Slot Attention}





\begin{icmlauthorlist}
\icmlauthor{Yanbo Wang}{TUDelft}
\icmlauthor{Letao Liu}{NTU}
\icmlauthor{Justin Dauwels}{TUDelft}
\end{icmlauthorlist}

\icmlaffiliation{TUDelft}{Department of EEMCS, Delft University of Technology, Delft, Netherlands}
\icmlaffiliation{NTU}{School of EEE, Nanyang Technological University, Singapore}

\icmlcorrespondingauthor{Yanbo Wang}{y.wang-27@tudelft.nl}

\icmlkeywords{Machine Learning, ICML}

\vskip 0.3in
]



\printAffiliationsAndNotice{}  

\begin{abstract}
Slot attention has shown remarkable object-centric representation learning performance in computer vision tasks without requiring any supervision. Despite its object-centric binding ability brought by compositional modelling, as a deterministic module, slot attention lacks the ability to generate novel scenes. In this paper, we propose the Slot-VAE, a generative model that integrates slot attention with the hierarchical VAE framework for object-centric structured scene generation. For each image, the model simultaneously infers a global scene representation to capture high-level scene structure and object-centric slot representations to embed individual object components. During generation, slot representations are generated from the global scene representation to ensure coherent scene structures. Our extensive evaluation of the scene generation ability indicates that Slot-VAE outperforms slot representation-based generative baselines in terms of sample quality and scene structure accuracy.


\end{abstract}

\section{Introduction}
\label{submission}

Human intelligence is capable of visually segmenting objects out of natural scenes, implicitly learning abstract object concepts, and creatively imagining novel scenes \citep{yuille2006vision} \citep{frankland2020concepts}. Equipping machines with such capabilities in an unsupervised way has been a desideratum for a long time \citep{johnson1983mental} \citep{ha2018world} \citep{wu2021greedy} \citep{scholkopf2021toward}, since this can facilitate intelligent agents understanding scenes, reasoning about object relationships, and performing tasks efficiently \citep{battaglia2013simulation} \citep{lake2017building} \citep{geiger2012we} \citep{cordts2016cityscapes} \citep{santoro2017simple} \citep{devin2018deep} \citep{greff2020binding} \citep{mambelli2022compositional}. To that end, most of the recent models resort to the variational autoencoder (VAE) framework \citep{kingma2013auto} \citep{rezende2014stochastic} for the purpose of joint object-centric representation inference and image generation. Depending on how to model the compositionality of images, existing works can be roughly categorized as spatial attention-based generative models and scene-mixture-based generative models. 

Spatial attention-based generative models infer object-centric representations by extracting a bounding box for each individual object \citep{eslami2016attend} \citep{crawford2019spatially} \citep{lin2020space} \citep{jiang2019scalor} \citep{jiang2020generative}. Such bounding boxes explicitly represent the position and size of object components enabling interpretable object manipulation. However, this type of model was pointed out to struggle to segment objects with extensively varied scales because the size of objects is, to some extent, presumed \citep{engelcke2021genesis} \citep{emami2022slot}. Moreover, rectangular bounding boxes are also not flexible enough to model image components of complex morphology \citep{lin2020space}.
In contrast, scene-mixture generative models decompose a visual scene into image-sized components (also known as slots), and infer slot representations corresponding to individual objects \citep{burgess2019monet} \citep{greff2019multi} \citep{engelcke2019genesis} \citep{engelcke2021genesis}. Such models segment objects with masks and are flexible enough to capture complex object components. Recent advances in scene-mixture models have shown remarkable object segmentation performance \citep{engelcke2019genesis} \citep{engelcke2021genesis}. However, although the design of such models advocates autoregressive priors for the purpose of generating coherent scenes, they are still unable to model object relationships in highly structured images and the generated samples are very blurry.


In this paper, we propose an object-centric generative model termed Slot-VAE that integrates slot attention with the hierarchical VAE framework for joint slot representation inference and structured image generation. In the proposed model, object-centric representation inference is achieved with the slot attention module \citep{locatello2020object}. Although slot attention has shown very impressive unsupervised segmentation performance, it is a deterministic module without the ability to generate novel scenes. If we naïvely combine slot attention with vanilla VAE for multi-object image generation, the generated images would be unreasonable because slots are completely independent and the scene structure (e.g., object relationships) is ignored. To overcome this issue, we adopt a two-layer hierarchical VAE model, which provides both global scene representations that capture the scene structure and object-centric slot representations that characterize individual objects. Slot representations are generated from global scene representations during the generation stage to ensure coherent scene structure. During training, besides learning from global scene representations, slot representations are also regularized by an independent prior to encourage object-centric disentanglement. Furthermore, the variational framework and independent prior also bring slot attention the attribute-level disentanglement. Evaluating on several multi-object datasets, we show that Slot-VAE outperforms baselines in terms of sample quality and scene structure learning.
 

The contributions of the paper are as follows. First, we introduce a generative model that embeds slot attention into the principled latent variable modelling framework for novel scene generation. Second, we incorporate a hierarchical latent variable model to learn both scene-level and object-centric representations. Third, we empower the slot attention baseline with object attribute-level disentanglement ability. Lastly, extensive experimental results suggest our proposed method outperforms the state-of-the-art methods in terms of sample quality and scene structure accuracy. 

\section{Related Works}

\textbf{Object-Centric Generative Modelling.} Compositional image modelling approaches \citep{greff2017neural} \citep{greff2017neural} \citep{kosiorek2018sequential} \citep{crawford2019spatially} \citep{burgess2019monet} \citep{greff2019multi}\citep{lin2020space} \citep{locatello2020object}  \citep{emami2021efficient} \citep{singh2021illiterate} \citep{kipf2021conditional}  \cite{seitzer2022bridging} \citep{https://doi.org/10.48550/arxiv.2211.01177} \citep{elsayed2022savi++} typically incorporate object locality as inductive bias or exploit simple decoder networks as reconstruction bottlenecks \citep{engelcke2020reconstruction} to achieve object-centric disentanglement. However, these approaches, unlike ours, cannot generate coherent novel scenes. GENESIS and GENESIS-V2 \citep{engelcke2019genesis} \citep{engelcke2021genesis} adopt autoregressive prior for coherent scene generation, but unlike ours, they lack the scene-level representation learning ability and generate blurry samples. GNM \citep{jiang2020generative} and similarly GSGN \citep{deng2021generative} resort to a hierarchical VAE model for both distributed and symbolic representations learning, but the bounding box representations therein prevent them from modelling complex objects or backgrounds, unlike ours where more flexible slot representations are used. SRI \cite{emami2022slot} learns slot representations and scene-level representations, but it has to sequentially infer object representations due to the assumed autoregressive posterior. In contrast, our approach poses an independent prior on slot representations allowing parallel inference. Besides, our approach trains the model without the need to learn a fixed object order, but SRI requires specialized auxiliary loss for object order learning so as to train the model.

\textbf{GANs for Compositional Generation:} GANs-based methods \citep{van2020investigating} \citep{nguyen2020blockgan} \citep{liao2020towards} \citep{niemeyer2021giraffe} \citep{ehrhardt2020relate} are able to map independent random noise vectors to individual object components on images allowing object-level controllability, but these models lack an inference process and thus cannot edit a given image unlike ours. Meanwhile, these GANs models share common unstable training issues.

\begin{figure*}
\centering
\includegraphics[width=13cm, height=6.5cm]{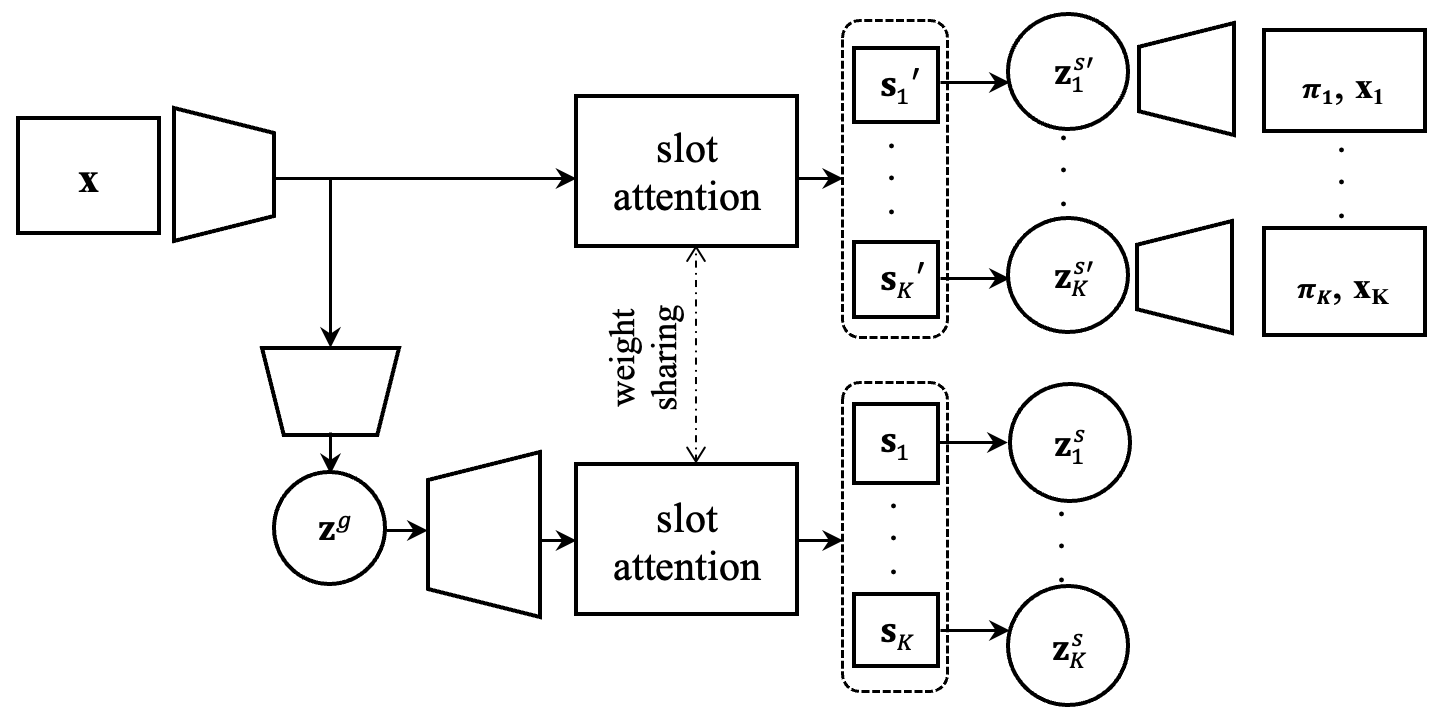}
\caption{Slot-VAE overview. The image $\mathbf{x}$ is passed through a CNN module. The obtained image features go through two paths in parallel. On the first path, the obtained image features are input into a slot attention module to learn slot representations $\{\mathbf{s}_{k}'\}_{k=1}^K$. From slots $\{\mathbf{s}_k'\}_{k=1}^K$, latent vectors $\{\mathbf{z}_k^{s'}\}_{k=1}^K$ are inferred. Then, a shared decoder decodes the individual object latent vector $\{\mathbf{z}_k^{s'}\}_{k=1}^K$ into object masks $\boldsymbol{\pi}_{1:K}$ and object components $\mathbf{x}_{1:K}$. By combining $\mathbf{x}_{1:K}$ with $\boldsymbol{\pi}_{1:K}$, the input $\textbf{x}$ is reconstructed. On the second path, the obtained image features is encoded into a global latent vector $\textbf{z}^g$. From $\textbf{z}^g$, a feature map is built and fed into a slot attention module to generate slot representations $\{\mathbf{s}_k\}_{k=1}^K $. From $\{\mathbf{s}_k\}_{k=1}^K $, latent vectors $\{\mathbf{z}_k^{s}\}_{k=1}^K$ are inferred. The two paths use the same slot attention module and share weights and initialization values, and it requires $\{\mathbf{z}_k^{s'}\}_{k=1}^K$ and  $\{\mathbf{z}_k^{s}\}_{k=1}^K$ to be as close as possible during training measured with KL divergence.} 
\label{Model_Overview}
\end{figure*}

\section{The Proposed Model: Slot-VAE}
\label{headings}
The overview of Slot-VAE is illustrated in Fig. \ref{Model_Overview}. 

\subsection{Generation}
For an image $\mathbf{x} \in [0, 1]^{H \times W \times C}$ with height $H$, width $W$ and $C$ channels, we postulate a two-layer hierarchical latent model for the potential image generation process. Specifically, the first-layer latent vector $\mathbf{z}^g \in \mathbb{R}^{L\times 1}$ captures the global structure in the image, for the purpose of modelling relationships among objects. Generated from $\mathbf{z}^g$, the second-layer latent vectors $\{\mathbf{z}_{k}^{s} \in \mathbb{R}^{D\times 1}\}_{k=1}^K$ represent each individual object in the image, with the goal of incorporating object-centric slot representations. These slot representations $\mathbf{z}_{1:K}^{s}$ are assumed to be conditionally independent given $\mathbf{z}^g$. Finally, with $\mathbf{z}_{1:K}^{s}$, an image $\mathbf{x}$ can be rendered with a decoder. Mathematically, the complete generative model can be written as:
\begin{equation}
    p_\theta(\mathbf{x}) = \iint p_\theta(\mathbf{x}\mid\mathbf{z}_{1:K}^{s}) p_\theta(\mathbf{z}_{1:K}^{s} \mid \mathbf{z}^g)p_\theta(\mathbf{z}^g)d\mathbf{z}_{1:K}^{s}d\mathbf{z}^g.
    \label{Complete_Generative_Model}
\end{equation}

The global latent vector $\mathbf{z}^g$ serves as an information bottleneck to extract high-level information (e.g., object appearance, positions and relations) for whole image reconstruction. $\mathbf{z}^g$ is similar to the latent vector in VAE but not exactly the same. The difference is that in VAE the latent vector is directly decoded to an image, while in Slot-VAE $\mathbf{z}^g$ is used to generate slot representations $\mathbf{z}_{1:K}^{s}$. For the prior of $\mathbf{z}^g$, we can choose a powerful StructDRAW prior \cite{jiang2020generative} or a simple Normal distribution depending on image complexity.

Slot representations $\mathbf{z}_{1:K}^{s}$, in contrast to $\mathbf{z}^g$, ideally embed information of individual object components and totally ignores object relationships. Such object-centric representations explicitly model the compositional structure of images, enable compositional generation and make the generation process interpretable. To generate $\mathbf{z}_{1:K}^{s}$ from $\mathbf{z}^g$, we first construct a feature map $\textbf{f}\in \mathbb{R}^{H\times W\times D}$ from $\mathbf{z}^g$ and then feed $\textbf{f}$ to a slot attention module \cite{locatello2020object} to obtain slot representations $\{\mathbf{s}_k\in \mathbb{R}^{D\times 1}\}_{k=1}^K $. Since slot attention is a deterministic module, an additional MLP is needed to map deterministic $\mathbf{s}_{1:K}$ to probabilistic latent vectors $\mathbf{z}_{1:K}^{s}$. Assuming $\mathbf{z}_{1:K}^{s}$ are Gaussian and
conditionally independent given $\mathbf{z}_g$, we have:
\begin{equation}
    p_\theta(\mathbf{z}_{1:K}^{s} \mid \mathbf{z}^g) = \prod_{k=1}^Kp_\theta(\mathbf{z}_{k}^{s} \mid \mathbf{z}^g).
    \label{Prior}
\end{equation}
The use of the slot attention module for object-centric latent vector generation sets the proposed Slot-VAE apart from GNM \citep{jiang2020generative} where bounding box extraction is adopted. Such a difference brings the following key benefits. First, slot-based models have been shown to be more flexible in modelling objects with complex morphology compared with the spatial attention module \citep{lin2020space}. Second, the dimension of the feature map $\textbf{f}$ in GNM fundamentally limits the maximum number of components in an image to be $H\times W$. Once a GNM model is trained, it at most can infer $H\times W$ objects. In contrast, the slot attention module can successfully generalize to infer more object components even though it only saw $K$ object components during training. Comes with these benefits a key challenge to Slot-VAE: there is no fixed order for the slot attention outputs. Since slot attention maps an input into a set (of slots), for the same input image, multiple runs may give the same set of slot representations but with different orders. This is because slot attention employs random initialization for slots to achieve slot permutation symmetry. However, such randomness makes the learning of a hierarchical latent variable model extremely challenging, which we will explain in detail in Section 3.3 and contribute to solving it.


With $\mathbf{z}_{1:K}^{s}$, rendering an image $\mathbf{x}$ is as follows. First, from $\mathbf{z}_{1:K}^{s}$ (or $\mathbf{z}_{1:K}^{s'}$ in Fig. \ref{Model_Overview}), $K$ sub-images $\{\mathbf{x}_{k} \in [0, 1]^{H \times W \times C}\}_{k=1}^K$ are rendered, each of which has the same dimension as $\mathbf{x}$ and ideally contains only one object. Meanwhile, this process also produces $K$ object masks $\boldsymbol{\pi}_{1:K} \in [0, 1]^{H \times W}$ corresponding to each $\mathbf{x}_{k}$. Then the image $\mathbf{x}$ is obtained by combining $\mathbf{x}_{1:K}$ with masks $\boldsymbol{\pi}_{1:K}$. Pixel-wisely, the likelihood can be written as 
\begin{equation}
    p_\theta(\mathbf{x}_{i, j} \mid \mathbf{z}_{1:K}^{s}) =  \mathcal{N}\bigg(\big(\sum_{k=1}^K\pi_{i, j, k}(\mathbf{z}_{1:K}^{s}) \mu_{i, j, k}(\mathbf{z}_k^{s})\big), \mathbf{\sigma}_x^2\bigg),
    \label{Likelihood}
\end{equation}
where $(i, j) $ is the pixel coordinate, $\mathbf{\sigma}_x$ is the standard deviation with a fixed value, and $\pi_{i, j, k}(\cdot)$ and $\mu_{i, j, k}(\cdot)$ are nonlinear functions mapping from latent vectors to masks $\boldsymbol{\pi}_{k}$ and mean values of $\mathbf{x}_{k}$ at pixel $(i, j)$. These nonlinear functions are parameterized by neural networks with learnable parameters $\theta$, and implementation details are provided in the appendix. In equation \ref{Likelihood}, $\pi_{i, j, k}$ serves as mixing probability, so it is constrained by $\sum_{k=1}^{K}\pi_{i, j, k}=1, \forall (i, j)$. 

In summary, to generate a novel scene, we first draw a random sample from the prior distribution of the global latent vector $\mathbf{z}^g$, from which a feature map $\textbf{f}$ is built. Then, object-centric latent vectors $\mathbf{z}_{1:K}^{s}$ are generated by using the slot attention module with the feature map $\textbf{f}$ as input. Finally, object components $\mathbf{x}_{1:K}$ and corresponding masks $\boldsymbol{\pi}_{1:K}$ are generated from $\mathbf{z}_{1:K}^{s}$ with parallel decoders, and a novel scene is rendered by combining $\mathbf{x}_{1:K}$ with $\boldsymbol{\pi}_{1:K}$.


\subsection{Inference}
Considering that the true posterior is intractable, we approximate the posterior with:
\begin{equation}
    p_\theta(\mathbf{z}^g, \mathbf{z}_{1:K}^{s} \mid \mathbf{x}) \approx  q_\phi(\mathbf{z}^{g} \mid \mathbf{x})q_\phi(\mathbf{z}_{1:K}^{s} \mid \mathbf{x}),
    \label{Posterior}
\end{equation}
wherein the global latent posterior $q_\phi(\mathbf{z}^{g} \mid \mathbf{x})$ is modelled by an autoregressive model or Gaussian distribution depending on StructDRAW prior or Gaussian prior is used \cite{jiang2020generative}.

We further assume the factorization $q_\phi(\mathbf{z}_{1:K}^{s} \mid \mathbf{x}) = \prod_{k=1}^Kq_\phi(\mathbf{z}_{k}^{s} \mid \mathbf{x})$. Such conditional independence assumption on the posterior distribution of slot representations enables the inference of individual $\mathbf{z}_{k}^{s}$ to be performed in parallel, which avoids sequential inference like in GENESIS. 
We adopt slot attention \citep{locatello2020object} followed by an MLP to infer $\mathbf{z}_{1:K}^{s}$, which is detailed as follows. 

\textbf{CNN for feature extraction.} Instead of directly working in the pixel domain, the slot representation inference starts from passing the input image $\mathbf{x}$ through a CNN backbone to extract a feature map $\textbf{f}_x=f_{enc}(\textbf{x}) \in \mathbb{R}^{H\times W\times D}$, where the CNN backbone is augmented with positional embeddings.

\textbf{Slot attention for component discovery.} To discover object components, the feature map $\textbf{f}_x$ is first flattened into vectors $\mathbf{f}_{input} \in \mathbb{R}^{{(H \times W)}\times D}$. Then, $\mathbf{f}_{input}$ is mapped to $K$ object slots $\mathbf{s}_{1:K}$ with a slot attention module.

\textbf{MLP for latent vector inference.} From slots $\mathbf{s}_{1:K}$, we would like to infer the latent variables $\mathbf{z}_{1:K}^s$. We assume the approximate posterior distribution of each individual slot $q_\phi(\mathbf{z}_{k}^{s} \mid \mathbf{x})$ to be Gaussian. Hence, inferring $\mathbf{z}_{k}^{s}$ is equivalent to infer Gaussian parameters $\{(\mathbf{\mu}_k^{s}$, $\mathbf{\sigma}_k^{s})\}_{k=1}^K$. To that end, we use an MLP shared across objects mapping from slots to Gaussian means and variances: $(\mathbf{\mu}_k^{s}, \mathbf{\sigma}_k^{s}) := \text{MLP}(\mathbf{s}_k)$.


\subsection{Training}

Given the above generative and inference model, the ELBO can be derived as follows:
\begin{equation}
\begin{aligned}
    \mathcal{L}(\mathbf{x}; \theta, \phi)& = \mathbb{E}_{q_\phi(\mathbf{z}_{1:K}^{s} \mid \mathbf{x})}\big[\text{log} p_\theta(\mathbf{x} \mid \mathbf{z}_{1:K}^{s})\big]\\
    &\quad -D_{\text{KL}}\big[q_\phi(\mathbf{z}_{1:K}^{s}\mid\mathbf{x}) \mid\mid p_\theta(\mathbf{z}_{1:K}^{s}\mid\mathbf{z}^{g})\big]\\
    &\quad\quad\quad- D_{\text{KL}}\big[q_\phi(\mathbf{z}^{g} \mid \mathbf{x}) \mid\mid p_\theta(\mathbf{z}^{g})\big]\\
\end{aligned}
\label{ELBO}
\end{equation}
where $D_{\text{KL}}(q || p)$ is Kullback-Leibler Divergence.


\textbf{Slot Order Matching in KL.} Observing the second term on the RHS of equation \ref{ELBO}, we can identify a key challenge for the calculation of this KL divergence: since the slots given by slot attention come with no fixed order, how can we determine the correspondence between $\mathbf{z}_{1:K}^{s}$ inferred from input $\textbf{x}$ (which is denoted $\mathbf{z}_{1:K}^{s'}$ in Fig. \ref{Model_Overview}) and $\mathbf{z}_{1:K}^{s}$ generated from $\mathbf{z}^{g}$? This challenge does not appear in GNM because the spatial attention module therein provides fixed order for each object component, which makes the calculation of KL divergence in GNM possible. To address such a challenge in Slot-VAE, we propose to implement $q_\phi(\mathbf{z}_k^{s} \mid \mathbf{x})$ and $p_\theta(\mathbf{z}_{k}^{s}\mid\mathbf{z}^{g})$ with a shared slot attention module. That is to say, as shown in Fig. \ref{Model_Overview}, the two slot attention modules share parameters. Meanwhile, slots $\textbf{s}_k^{'}$ and $\textbf{s}_k$ in Fig. \ref{Model_Overview} share initialization values. Intuitively, such an architecture design encourages the feature map $\textbf{f}$ generated from $\textbf{z}_g$ to be consistent with the feature map $\textbf{f}_x$ encoded from input $\textbf{x}$. With similar inputs and the same random initialization values, we can expect the output of the two slot attention modules could keep close to each other. As a result, the order of $\textbf{s}_k$ (or $\mathbf{z}_k^s$) can have a good chance to align well with that of $\textbf{s}_k^{'}$ (or $\mathbf{z}_k^{s'}$) in Fig. \ref{Model_Overview}, enabling the calculation of $D_{\text{KL}}\big[q_\phi(\mathbf{z}_{1:K}^{s}\mid\mathbf{x}) \mid\mid p_\theta(\mathbf{z}_{1:K}^{s}\mid\mathbf{z}^{g})\big]$. We will empirically demonstrate the efficacy of such an architectural inductive bias for slot order matching in Section 4.

Furthermore, since $p_\theta(\mathbf{z}_{1:K}^{s}\mid\mathbf{z}^{g})$ in the second term of equation \ref{ELBO} is learned from the posterior distribution $p_\theta(\mathbf{z}_g\mid\mathbf{x})$, it provides no explicit prior information to guide the learning of the posterior distribution $q_\phi(\mathbf{z}_{1:K}^{s}\mid\mathbf{x})$ during training. To explicitly provide guidance to the learning of $q_\phi(\mathbf{z}_{1:K}^{s}\mid\mathbf{x})$, the following auxiliary loss could be incorporated:
\begin{equation}
    \mathcal{L}_{aux} = -D_{\text{KL}}\big[q_\phi(\mathbf{z}_{1:K}^{s}\mid\mathbf{x}) \mid\mid \prod_{k=1}^K\mathcal{N}(\mathbf{0}, \mathbf{1})\big],
\label{auxiliary loss}
\end{equation}
where independent normal prior constrains $\mathbf{z}_{1:K}^{s}$ to be independent on each other. As a result, such independence encourages each slot representation to capture only a single object leading to object-centric disentanglement. Meanwhile, attribute-level disentanglement within an object can also be achieved due to diagonal variance of the normal prior, which we will show in experiments. 

Combining the derived ELBO in equation \ref{ELBO} and the introduced auxiliary loss in equation \ref{auxiliary loss}, the overall objective function is:
\begin{equation}
    \tilde{\mathcal{L}} = \mathcal{L} + \mathcal{L}_{aux},
\end{equation}
which is minimized to train Slot-VAE. For effective training, we also introduce hyperparameters to balance the reconstruction loss and KL terms \citep{rezende2018taming} \citep{fu2019cyclical}, which will be detailed in the Appendix.



\section{Experiments}
\label{others}
The experiments are to evaluate: i) image decomposition performance, ii) sample quality and structure accuracy of generated samples, iii) and disentanglement performance. 
\begin{figure*}[h!]
\center
\includegraphics[width=15cm, height=7cm]{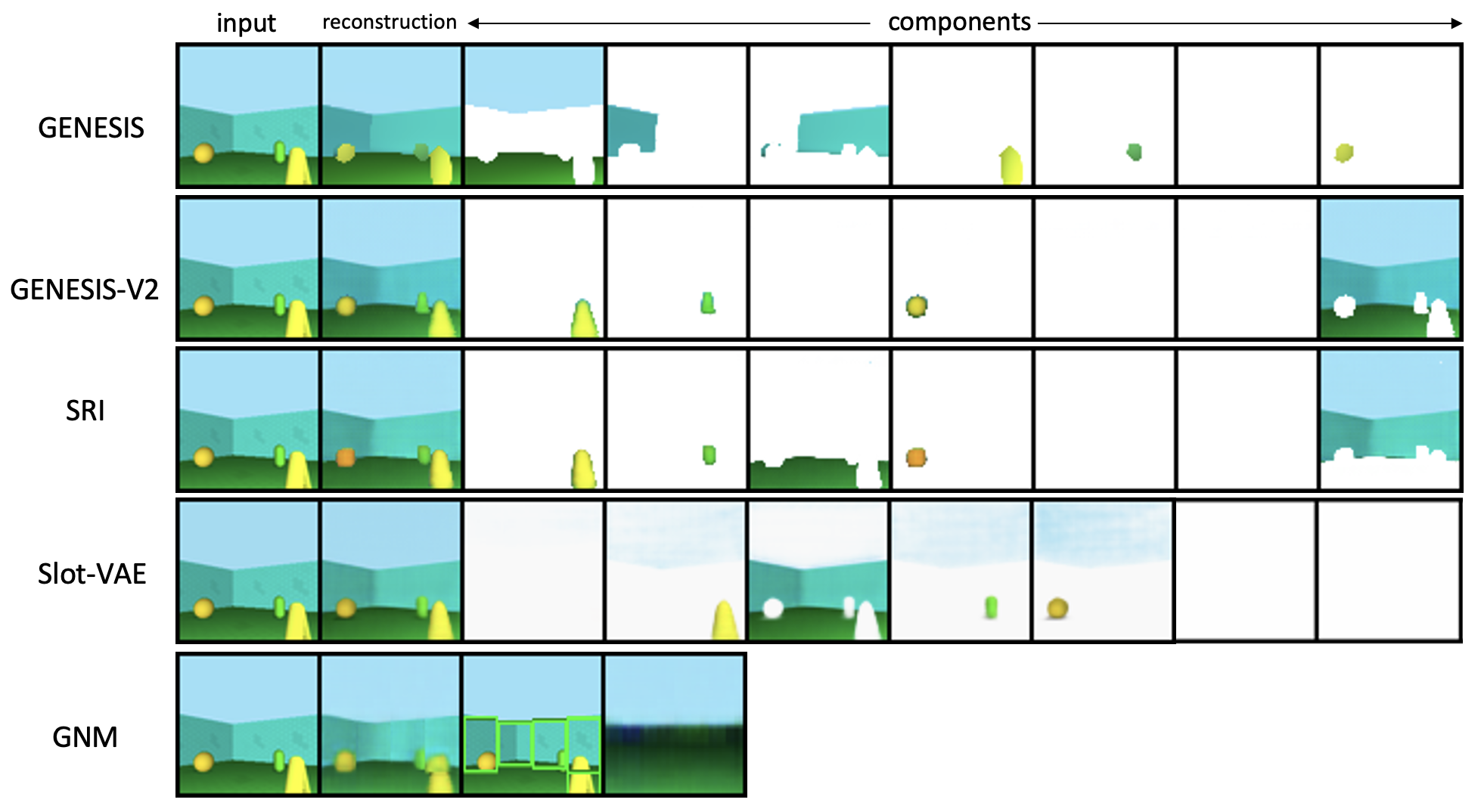}
\caption{Image decompostion and reconstruction performance on the ObjectsRoom dataset.}
\label{ObjectRoom_decom}
\end{figure*}

\textbf{Dataset.} The experiments involve three datasets including \emph{ObjectRoom} \citep{multiobjectdatasets19}, \emph{ShapeStacks} \citep{groth2018shapestacks} and \emph{Arrow Room}\citep{jiang2020generative}. The datasets \emph{ObjectRoom} and \emph{ShapeStacks} are commonly used by previous works to test object-centric inference and generation, while \emph{Arrow Room} is less considered because this dataset is highly structured and its probabilistic density is hard to model. 
In \emph{Arrow Room}, there is always an arrow shape object in the front of the scene and three objects in the back. Two of the three objects in the back have the same shape, while a third one has a unique shape. The arrow in the front always points to the object with a unique shape in the back.

\textbf{Baselines.} We compare Slot-VAE against state-of-the-art object-centric generative models including GENESIS, GENESIS-V2, SRI and GNM. In these baselines, GNM is based on the spatial attention model (i.e., bounding box representations) with hierarchical generation process, while GENESIS, GENESIS-V2 and SRI are scene-mixture models (i.e., slot representations) that assume an autoregressive prior. Some of the baseline models already released their trained models for \emph{ObjectRoom}, \emph{ShapeStacks} or \emph{Arrow Room}. We do not retrain them and directly use their weights for comparison. For these of baseline models without trained models on some datasets, we train them with the official code.

\subsection{Qualitative Comparison of Image Decomposition, Image Reconstruction and Sample Generation}

\begin{figure*}[h!]
\center
\includegraphics[width=15cm, height=6cm]{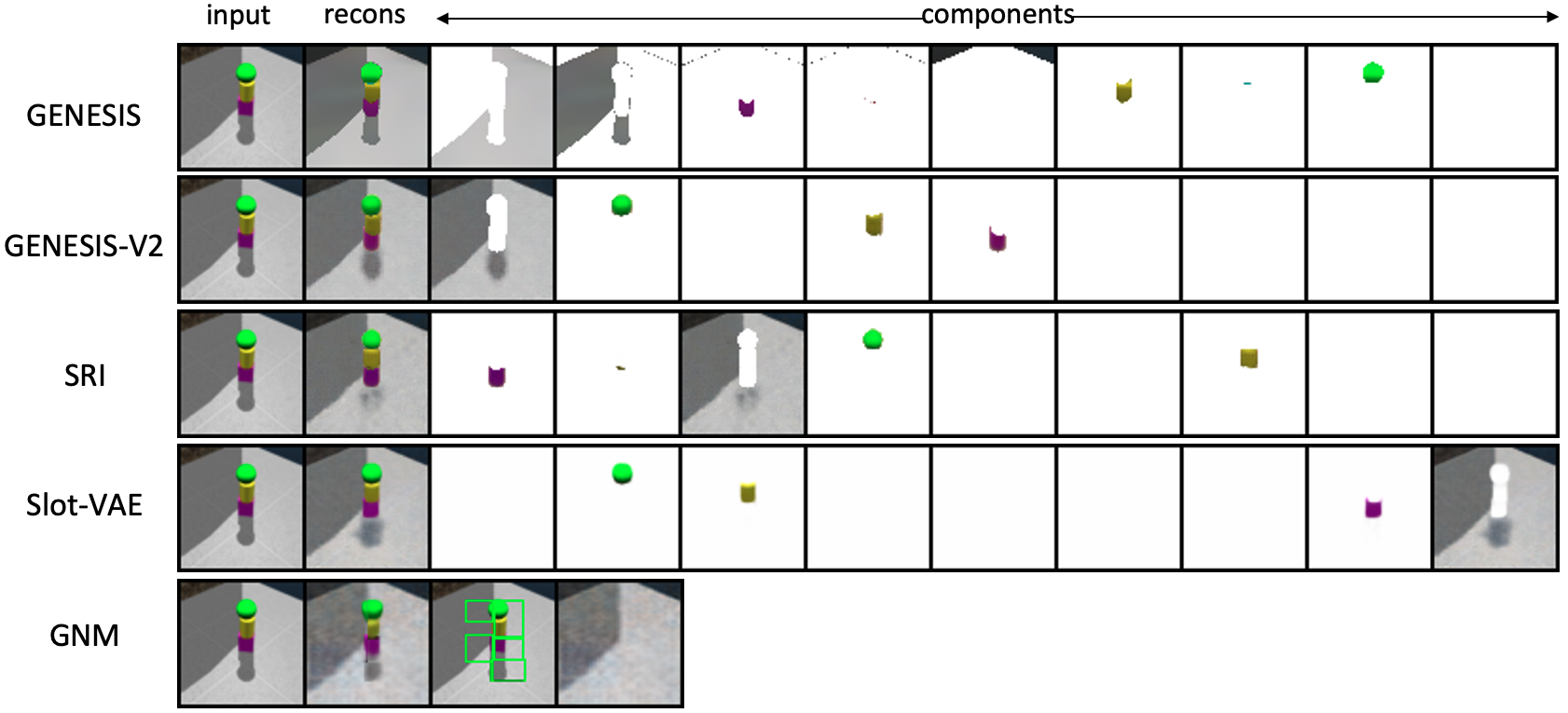}
\caption{Image decompostion and reconstruction performance on the ShapeStacks dataset.}
\label{ShapeStacks_decom}
\end{figure*}

\begin{figure*}[h!]
\center
\includegraphics[width=12cm, height=7cm]{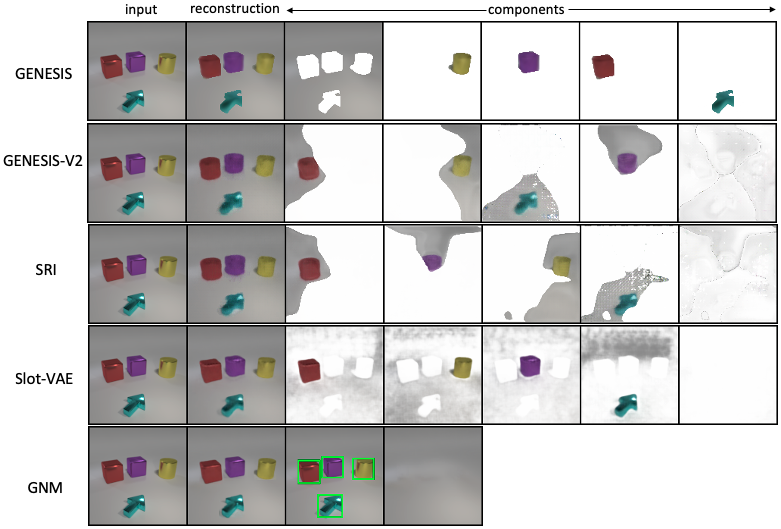}
\caption{Image decompostion and reconstruction performance on the Arrow Room dataset.}
\label{ImageReconstruction}
\end{figure*}

\textbf{Decomposition and Reconstruction Performance}. We illustrate the input, reconstruction and decomposed object components of Slot-VAE and baselines in Fig. \ref{ObjectRoom_decom} - \ref{ImageReconstruction}. Note that GNM infers bounding box representations instead of slot representations. So in the figures, GNM has only two components, one for the foreground with bounding boxes and another for the background.

As shown in Fig. \ref{ObjectRoom_decom}, for the \emph{ObjectRoom} dataset that comes with simple object shapes and complex background components, scene mixture models GENESIS, GENESIS-V2, SRI and Slot-VAE achieve comparable decomposition and reconstruction performance. The only difference is that some of them capture the background with one slot while others use multiple slots. In contrast, GNM fails to segment objects correctly. It segments the scene into stripes containing parts of objects and parts of the background, and a single object is segmented into multiple bounding boxes. As a result, the reconstructed images of GNM show rectangular artifacts and objects are blurred. This is not surprising because with the use of grid sampling and bounding box representations, spatial-attention generative models like GNM struggle with modelling objects that have complex morphology. In Fig. \ref{ShapeStacks_decom}, we observe similar results for the \emph{ShapeStacks} dataset, where  GENESIS, GENESIS-V2, SRI and Slot-VAE decompose and reconstruct the image reasonably well while GNM again tries to model one single object with multiple bounding boxes. Failing to learn correct obeject-centric representations, GNM will also suffer during the generation stage as will be presented below. For the \emph{Arrow Room} dataset that has simple object shapes but complex scene structures in Fig. \ref{ImageReconstruction}, we can see all models successfully segment objects out of the scene and reconstruct the input image. However, GENESIS-V2 and SRI learn object representations that severely involve part of the background. Such representations will make the generated image samples very blurry, as will be shown below. We conjecture this is because the \emph{Arrow Room} dataset has too strong object position relationships, and GENESIS-V2 and SRI (based on GENESIS-V2) do not have enough capacity and have to choose simple ways to segment images. In summary, Slot-VAE achieves either better or comparable segmentation and reconstruction performance in comparison to baselines. Additional decomposition results of Slot-VAE can be found in the Appendix.

\begin{figure*}[h]
\center
\includegraphics[width=17cm, height=10cm]{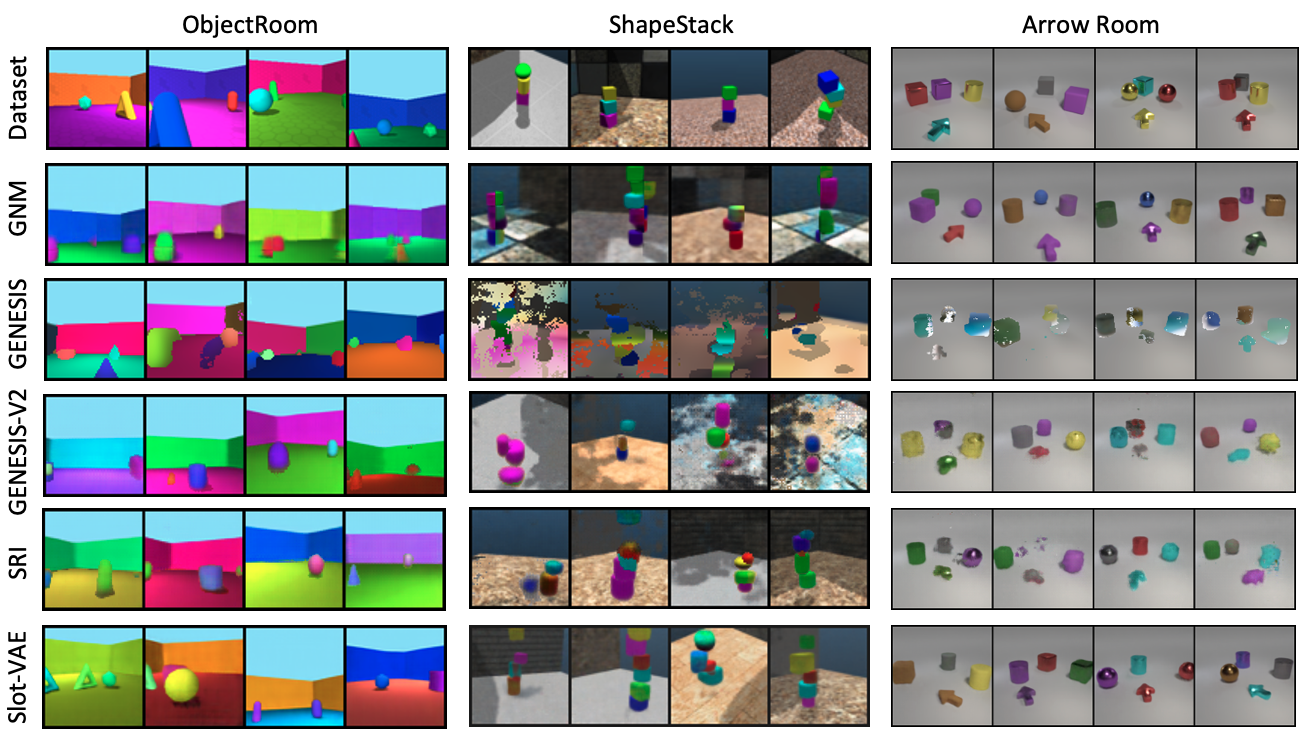}
\caption{Datasets and generation examples of Slot-VAE and baselines.}
\label{ImageGeneration}
\end{figure*}

\textbf{Generation Performance}. We show random samples generated by Slot-VAE and baseline models in Fig. \ref{ImageGeneration}. It can be seen Slot-VAE generates the sharpest samples that highly resemble all the datasets. For \emph{ObjectRoom}, samples generated by GNM show stripe artifacts due to its inaccurate object-centric representations captured by bounding boxes as discussed above. The sample quality of SRI is better than that of GENESIS and GENESIS-V2, but not as good as the proposed Slot-VAE. This can be reflected by the sharpness of object edges in the images. One can more easily identify object shapes (e.g., balls and triangles) with Slot-VAE compared to baselines. For \emph{ShapeStacks}, GNM again shows its limitation where it generates one individual object component with several parts. For example, a cube is represented by two small parts with completely different colors. Only SRI and Slot-VAE generate reasonable samples reflecting the scene structure of the \emph{ShapeStacks} dataset (i.e., one object is stacked on another), while the sample quality of Slot-VAE is better in terms of sharp object edges. For \emph{Arrow Room}, the most structured dataset, we find samples generated by GENESIS, GENESIS-V2 and SRI are very blurry and seldom show the underlying true scene structure (i.e., the arrow in the front always points to the object with a unique shape in the back). Both arrow directions or object shapes are not properly learned. This indicates that the autoregressive prior adopted in GENESIS, GENESIS-V2 and SRI is not strong enough to capture the complex scene structure in \emph{Arrow Room}. In contrast, GNM and Slot-VAE, both exploiting hierarchical model to capture scene structure, generate very coherent and high-quality samples on the \emph{Arrow Room} dataset. The reason why GNM works better on \emph{Arrow Room} in comparison to \emph{ObjectRoom} and \emph{ShapeStacks} is that object shapes are simple in \emph{Arrow Room}. In summary, Slot-VAE outperforms baselines in terms of sample quality and scene structure learning. Additional random generation results of Slot-VAE can be found in the Appendix.

\begin{table}[t]
\caption{ARI-FG ($\uparrow$) for Slot-VAE and Baselines on ObjectsRoom and ShapeStacks. Mean and standard deviation of ARI with three runs are presented. Scores labelled with ${}^*$ are from original works \citep{engelcke2020reconstruction} and \citep{emami2022slot}.}
\label{ARI-table}
\vskip 0.15in
\begin{center}
\begin{small}
\begin{sc}
\begin{tabular}{lcccr}
\toprule
Model & ObjectsRoom  & ShapeStacks\\
\midrule
GNM         & 0.63\textsuperscript{*} $\pm$ 0.00& 0.37\textsuperscript{*} $\pm$ 0.07\\
GENESIS     & 0.63\textsuperscript{*} $\pm$ 0.03 & 0.70\textsuperscript{*} $\pm$ 0.05\\
GENESIS-V2  & 0.84\textsuperscript{*} $\pm$ 0.01 & 0.81\textsuperscript{*} $\pm$ 0.00\\
SRI         & 0.83\textsuperscript{*} $\pm$ 0.02 & 0.78\textsuperscript{*} $\pm$ 0.02\\
Slot-VAE (ours) &  0.79 $\pm$ 0.01 & 0.80 $\pm$ 0.01\\
\bottomrule
\end{tabular}
\end{sc}
\end{small}
\end{center}
\vskip -0.1in
\end{table}

\begin{table*}
\caption{Fréchet Inception Distances (FID $\downarrow$) and Structure Accuracy (S-Acc $\uparrow$) for Slot-VAE and Baselines. Mean and standard deviation of FID with three runs are presented. Scores labelled with ${}^*$ are from original works \citep{engelcke2020reconstruction} and \citep{emami2022slot}.}
\label{FID-table}
\vskip 0.15in
\begin{center}
\begin{small}
\begin{sc}
  \begin{tabular}{@{\extracolsep{8pt}}lccccr@{}}
    \toprule
    \multirow{2}{*} &
      \multicolumn{1}{c}{ObjectsRoom} &
      \multicolumn{1}{c}{ShapeStacks} &
      \multicolumn{2}{c}{Arrow Room} \\ \cmidrule(l{12pt}r{10pt}){2-2} \cmidrule(l{12pt}r{10pt}){3-3} \cmidrule(l{12pt}r{10pt}){4-5}
      Model & {FID} & {FID} & {FID} & {S-Acc} \\
      \midrule
    GNM  & 51.6\textsuperscript{*}{$\pm$5}   & 49.3\textsuperscript{*}$\pm$2 & 11.2$\pm$2 & 0.97 \\
    \midrule
   GENESIS & 62.8\textsuperscript{*}$\pm$3  & 186.8\textsuperscript{*}$\pm$18 &  173.8$\pm$13& 0.11 \\
GENESIS-V2 & 52.6\textsuperscript{*}$\pm$3 & 112.7\textsuperscript{*}$\pm$3 & 111.8$\pm$5 & 0.20 \\
SRI        & 48.4\textsuperscript{*}$\pm$4 & 70.4\textsuperscript{*}$\pm$3 & 123.3$\pm$2 & 0.18 \\
Slot-VAE (ours)  & \textbf{34.9}$\pm$\textbf{1}  & \textbf{50.0} $\pm$ \textbf{1}  & \textbf{60.3}$\pm$\textbf{1} & \textbf{0.94} \\
    \bottomrule
  \end{tabular}
\end{sc}
\end{small}
\end{center}
\vskip -0.1in
\end{table*}

 \begin{figure*}[h]
\center
\includegraphics[width=15cm, height=8cm]{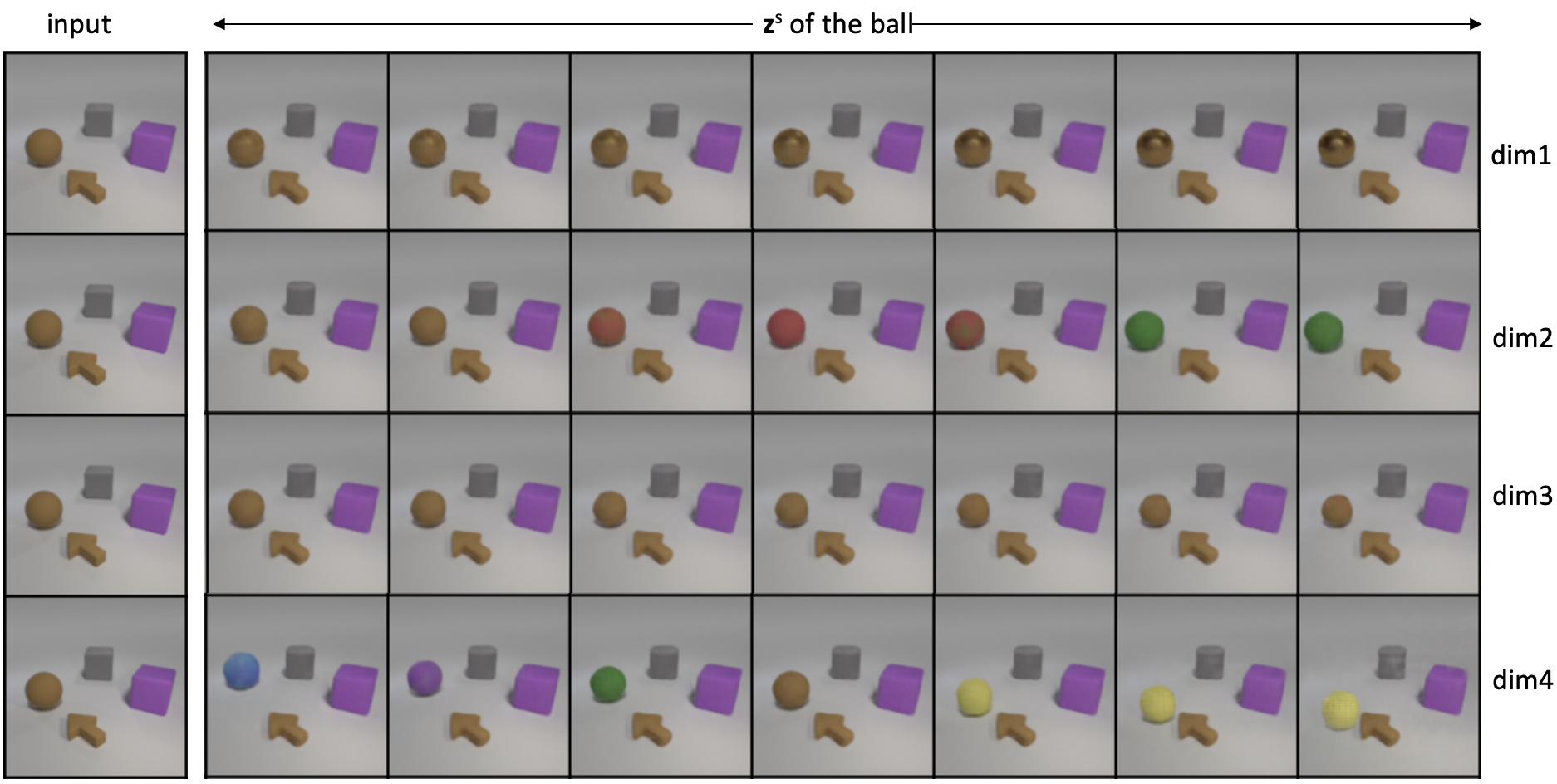}
\caption{Slot-VAE latent traversal on \emph{Arrow Room}. Each row only varies a certain dimension of $\mathbf{z}^s$ corresponding to the ball object.}
\label{LatentSpaceTraversalOurs}
\end{figure*}

\textbf{Scene Manipulation}. We elaborate on controllable scene generation to highlight the disentanglement performance of Slot-VAE. In Fig. \ref{LatentSpaceTraversalOurs}, in each row we vary a certain dimension of the object-centric latent vector corresponding to the ball object while keeping other object-centric latent vectors unchanged. As is shown, only attributes of the ball are changed in each row, and all other objects remain unaffected. Such object-level disentanglement is very useful for image editing and compositional generation. Besides object-level disentanglement, attributes-level disentanglement also naturally appears in Slot-VAE due to the adopted probabilistic framework. As shown in Fig. \ref{LatentSpaceTraversalOurs}, when we vary dimension 1, the texture of the ball changes; when we vary dimension 2, the color of the ball changes; when we vary dimension 3, the size of the ball changes. Although some dimensions (e.g., dim 4) entangle color and position a little, this can be further improved with existing attribute-level disentanglement techniques like $\beta$-VAE \citep{higgins2017beta} or $\beta$-TCVAE \citep{chen2018isolating}, which is out of the scope of this paper. In the proposed Slot-VAE, attribute-level disentanglement is a by-product brought by the VAE framework. By contrast, the original deterministic slot attention module comes with no obvious attribute-level disentanglement as analyzed in \citep{https://doi.org/10.48550/arxiv.2211.01177}. 
 

\subsection{Quantitative Comparison}
We report the Adjusted Rand Index (ARI) \citep{hubert1985comparing} score, Frechet Inception Distance (FID) \citep{heusel2017gans} score and scene structure accuracy (S-Acc) \citep{jiang2020generative} score to quantitatively evaluate the decomposition performance, sample quality, and scene structure accuracy. Since the \emph{Arrow Room} dataset comes with no ground truth masks, the ARI score on this dataset is not calculated. As shown in Table \ref{ARI-table}, slot-VAE achieves comparable ARI scores to baselines. For the FID score, the calculation involves 10000 real and generated samples. Table \ref{FID-table} reflects non-trivial FID score improvement by Slot-VAE against slot-representation baselines, highlighting the sample quality of Slot-VAE. Although the FID score of GNM on \emph{ObjetsRoom} and \emph{ShapeStacks} seems quite good, it should be emphasized that the generated images are unrealistic (i.e., generated objects are composed of multiple rectangular parts) due to inaccurate object representation learning as analyzed in the qualitative comparison results. For the S-Acc score, we manually classified 100 generated images per model, and calculated the ratio of successful images that correctly reflect scene structure. The datasets \emph{ObjetsRoom} and \emph{ShapeStacks} have relatively less clearly defined structures, which may result in difficulty in deciding if generated images truly reflect scene structures. To reduce subjective decisions, we mainly evaluate S-Acc of Slot-VAE and baseline models on the \emph{Arrow Room} dataset because this dataset has a clearly defined structure: the arrow object should always point to the object with a unique shape in the back. Slot-VAE achieves the best S-Acc score among all the slot representation-based models (GENESIS, GENESIS-V2 and SRI), as is shown in Table \ref{FID-table}.

\subsection{Ablation Study}
We further conduct experiments to demonstrate the efficacy of the proposed architectural design in Fig. \ref{Model_Overview}. Specifically, we aim to answer the following questions: (1) whether slot attention is necessary for generating slot representations from the global representation and (2) whether slot attention weight sharing and initialization value sharing are necessary for slot order matching. To that end, we evaluated the FID score and S-Acc score of several Slot-VAE variants.

\begin{table}[t]
\caption{FID ($\downarrow$) score and S-Acc ($\uparrow$) score of Slot-VAE and variants on the \emph{Arrow Room} dataset.}
\label{Ablation}
\vskip 0.15in
\begin{center}
\begin{small}
\begin{sc}
\begin{tabular}{lcccr}
\toprule
Model & FID   & S-Acc \\
\midrule
Slot-VAE        & \textbf{60.3±1} & \textbf{0.94}\\
Slot-VAE-MLP     & 289±9 & 0.00\\
Slot-VAE-Transformer  & 182.1±3 & 0.03\\
Slot-VAE-W/O-WS        & 215.5±2 & 0.00\\
Slot-VAE-W/O-IVS & 142.1±3 & 0.05\\
\bottomrule
\end{tabular}
\end{sc}
\end{small}
\end{center}
\vskip -0.1in
\end{table}

To answer question (1), we investigate two approaches that could be used as alternatives to slot attention to generating slot representations $\{\mathbf{s}_k\}_{k=1}^K$ from the global representation $\textbf{z}^g$. The first approach (termed as Slot-VAE-MLP) is by using an MLP to directly map the $\textbf{z}^g$ to $\{\mathbf{s}_k\}_{k=1}^K$. Although this approach is straightforward, it cannot work well intuitively. Specifically, an MLP learns a deterministic mapping that always outputs slots $\{\mathbf{s}_k\}_{k=1}^K$ with a fixed order for a given global latent vector, whereas the slots $\{\mathbf{s}_k'\}_{k=1}^K$ that are directly inferred from the input image with slot attention come with a random order. As a result, the order of $\{\mathbf{z}_k^{s}\}_{k=1}^K$ and that of $\{\mathbf{z}_k^{s'}\}_{k=1}^K$ can rarely match each other, leading to fluctuating KL divergence $D_{\text{KL}}\big[q_\phi(\mathbf{z}_{1:K}^{s}\mid\mathbf{x}) \mid\mid p_\theta(\mathbf{z}_{1:K}^{s}\mid\mathbf{z}^{g})\big]$ between slot prior and slot posterior and hence diverged training. This can be reflected by the very high FID score and low S-Acc score in Table \ref{Ablation}. The second approach (termed as Slot-VAE-Transformer) is by using a transformer to map the global vector $\textbf{z}^g$ and random initialization values of slots $\{\mathbf{s}_k\}_{k=1}^K$ shared with $\{\mathbf{s}_k'\}_{k=1}^K$ to slot representations. In this approach, slots generated by the transformer is permutation invariant due to random initialization, which addresses the fixed slot order issue in Slot-VAE-MLP. Intuitively, with shared initialized values, slots $\{\mathbf{s}_k\}_{k=1}^K$ generated from $\textbf{z}^g$ and slots $\{\mathbf{s}_k'\}_{k=1}^K$ inferred from the input image could have a good chance to match each other. Indeed, with this approach, our model matches the orders of the slots well. However, the generated slots turn out not so good in the sense that their corresponding decoded object components are very blurry. As a result, Slot-VAE-Transformer also has a very high FID score and a low S-Acc score. In contrast, Slot-VAE outperforms Slot-VAE-MLP and Slot-VAE-Transformer significantly, which demonstrates the effectiveness of slot attention for generating slot representations from the global representation.

To answer question (2), we trained a variant of Slot-VAE (termed as Slot-VAE-W/O-WS) without the weight sharing strategy in Fig. \ref{Model_Overview}. In this case, the two slot attention modules update their weights respectively with no common initialization values. Without weight sharing, we anticipate that the KL divergence $D_{\text{KL}}\big[q_\phi(\mathbf{z}_{1:K}^{s}\mid\mathbf{x}) \mid\mid p_\theta(\mathbf{z}_{1:K}^{s}\mid\mathbf{z}^{g})\big]$ could be large because the learned slot representations of the two attention modules can be quite different, which may result in unrealistic generation samples. This is demonstrated by the experimental results in Table \ref{Ablation}. We also trained another variant of Slot-VAE (termed as Slot-VAE-W/O-IVS) with weight sharing between the two slot attention modules but without initialization value sharing. Without initialization value sharing, the order of slots $\{\mathbf{s}_k\}_{k=1}^K$ generated from $\textbf{z}^g$ and the order of slots $\{\mathbf{s}_k'\}_{k=1}^K$ inferred from the input image cannot match each other very well. As a result, the KL divergence $D_{\text{KL}}\big[q_\phi(\mathbf{z}_{1:K}^{s}\mid\mathbf{x}) \mid\mid p_\theta(\mathbf{z}_{1:K}^{s}\mid\mathbf{z}^{g})\big]$ can not be properly calculated, and generated samples cannot reflect the dataset structure as quantitatively shown in Table \ref{Ablation}.

In summary, we empirically find that the slot attention module for generating slot representations from the global representation, weight sharing and initialization value sharing between the two attention modules  improve the generation performance significantly.


\section{Conclusion}
We propose an object-centric generative model, Slot-VAE, that integrates the slot attention module with a hierarchical VAE model for joint object-centric representation inference and scene structure modelling. The proposed model can discover object components in an unsupervised way and generate novel scenes controllable at both the object and attribute level. Experiment results show that Slot-VAE achieves better sampling quality and scene structure accuracy compared to slot representation-based generative baselines. 

One limitation of Slot-VAE is that the adopted slot attention module requires simple decoders like SBD \citep{watters2019spatial} to serve as a reconstruction bottleneck to decompose objects, which, however, may not scale to complex real-world scenes. This can be improved by using a transformer decoder \citep{singh2021illiterate} or diffusion model-based decoder \citep{jiang2023object}, which we leave for future work.

\section*{Social Impact}
The proposed Slot-VAE model shows no negative social impacts in its current form since the evaluation is carried out on synthetic datasets at this stage. However, with improved slot representation learning modules available in the future, our model has the potential to be applied to generate more sophisticated and realistic scenes. In that case, misuse should be avoided for malicious purposes. Proper use of the proposed model can actually benefit practical applications like artwork generation, scene understanding, and dataset augmentation, to name just a few.



\nocite{langley00}
\clearpage
\bibliography{example_paper}
\bibliographystyle{icml2023}

\newpage
\appendix
\onecolumn
\section{Additional Results of Slot-VAE.}

\begin{figure*}[h!]
\center
\includegraphics[width=10cm, height=8cm]{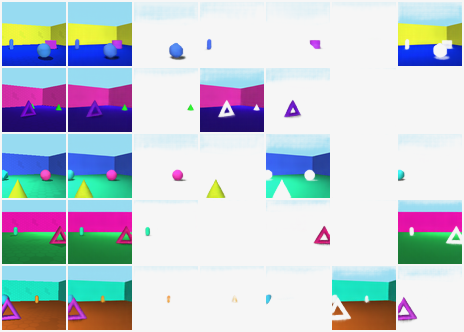}
\caption{Additional decomposition resulst of Slot-VAE (ObjectsRoom dataset).}
\label{Addi_DEC_Object}
\end{figure*}

\begin{figure*}[h!]
\center
\includegraphics[width=15cm, height=8cm]{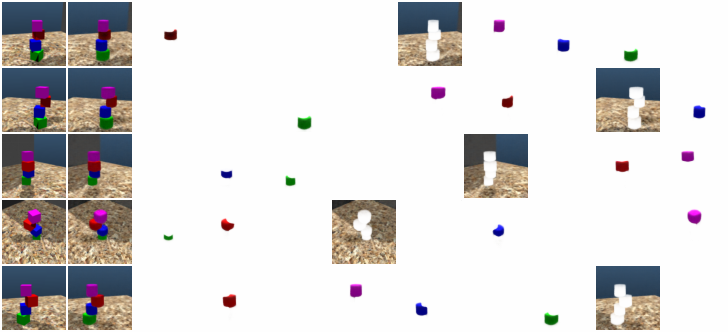}
\caption{Additional decomposition resulst of Slot-VAE (ShapeStacks dataset).}
\label{Addi_DEC_Shape}
\end{figure*}

\begin{figure*}[h!]
\center
\includegraphics[width=14cm, height=10cm]{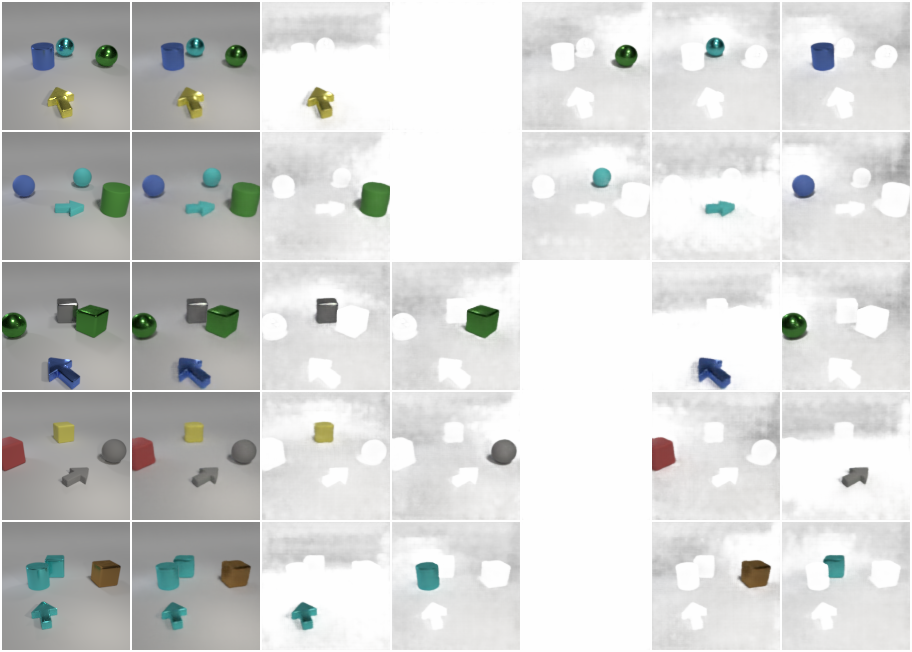}
\caption{Additional decomposition resulst of Slot-VAE (ShapeStacks dataset).}
\label{Addi_DEC_Arrow}
\end{figure*}

\begin{figure*}[h!]
\center
\includegraphics[width=16cm, height=8cm]{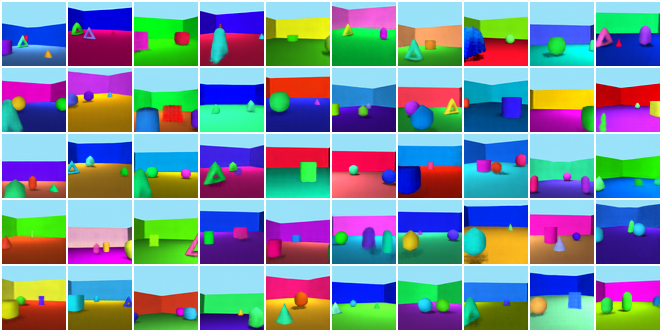}
\caption{Additional generation resulst of Slot-VAE (Arrow Room dataset).}
\label{Addi_GEN_Object}
\end{figure*}

\begin{figure*}[h!]
\center
\includegraphics[width=16cm, height=8cm]{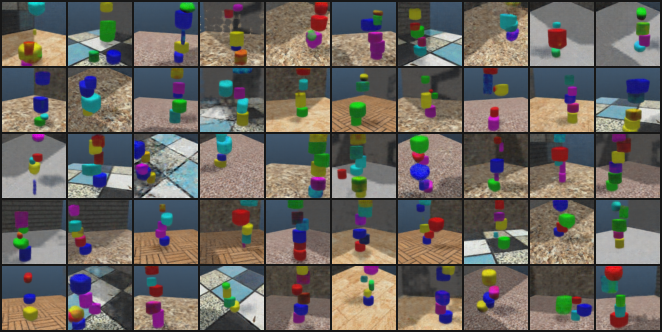}
\caption{Additional generation resulst of Slot-VAE (ShapeStacks dataset).}
\label{Addi_GEN_Shape}
\end{figure*}

\begin{figure*}[h!]
\center
\includegraphics[width=16cm, height=8cm]{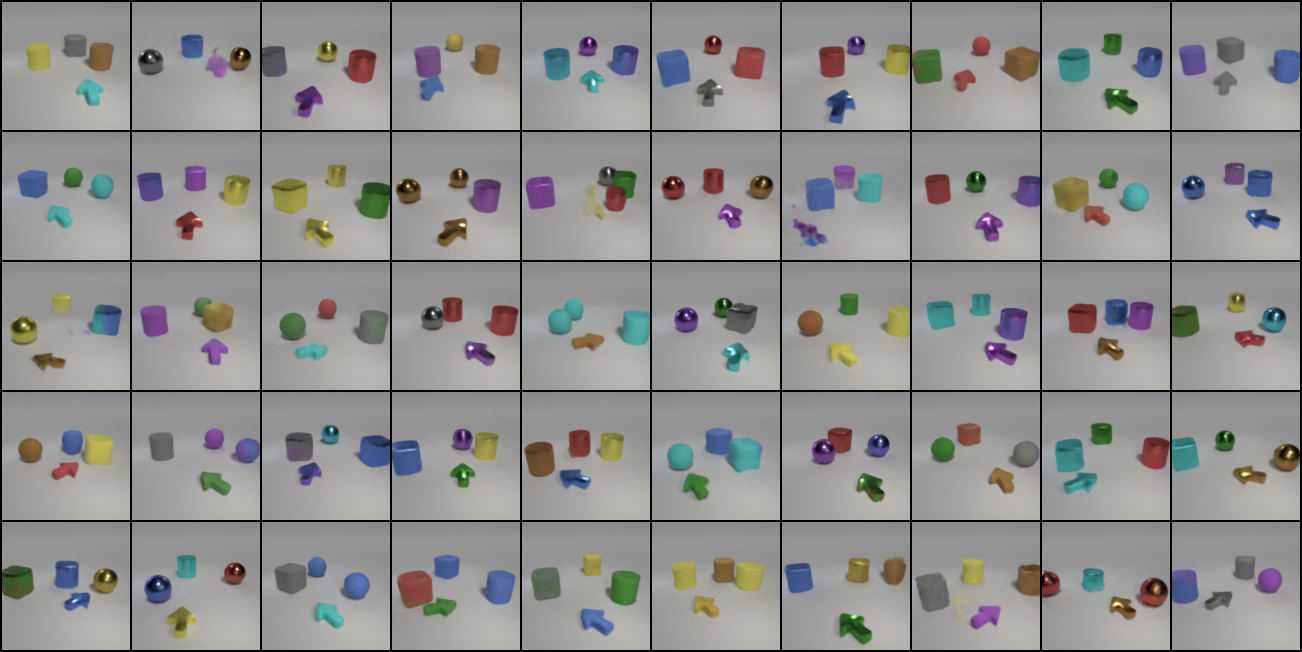}
\caption{Additional generation resulst of Slot-VAE (Arrow Room dataset).}
\label{Addi_GEN_Arrow}
\end{figure*}

We show additional scene decomposition and novel scene generation examples of Slot-VAE on \emph{ObjectsRoom} \emph{ShapeStacks} and \emph{Arrow Room} in Fig.\ref{Addi_DEC_Object} - Fig. \ref{Addi_GEN_Arrow}

\newpage
\section{Implementation Details of Slot-VAE.}
In this section, we introduce the implementation details of Slot-VAE. As shown in Fig. \ref{Model_Overview}, Slot-VAE has two parallel paths to train a two-layer hierarchical VAE model, which mainly includes the following four modules. 

\textbf{CNN backbone.} Before inferring the global latent representation and slot representations, the input image is first fed into a convolutional neural network to extract relatively high-level features. This convolutional neural network has 4 layers, each layer is with kernel size 5 and stride 1 and the final layer has 64 channels. The obtained feature map $\mathbf{f}_x$ still has image-sized dimensions and each feature (channel) has a dimension of 64, i.e., the dimension is $H\times W \times 64$. Soft position embedding are then added to the feature map to provide position information for the following modules. 

\textbf{Slot Attention Module.} On the first path, we adopt the slot attention module \citep{locatello2020object} for object-centric representation learning. We include the details for self-containing purpuse. To prepare for slot learning, the feature map $\mathbf{f}_x$ is first flattened into vectors $\mathbf{f}_{input}$ with dimension $(H\times W) \times 64$. To cluster the feature vectors into object components, the clustering center, i.e., slots, should be initialized first. The initialization values for object slots are from  Gaussian distribution respectively, i.e., $\mathbf{s}_{1:K} \sim \mathcal{N}(\mathbf{\mu}, \text{diag}(\mathbf{\sigma})) \in \mathbb{R}^{K\times 64}$, where $\mathbf{\mu}$ and $\mathbf{\sigma}$ are learnable parameters. These slots are then updated iteratively to compete for explaining feature vectors $\mathbf{f}_{input}$. The slot competition is achieved via a softmax-based attention mechanism :$\text{attn}_{i, j} := \frac{\text{exp}(M_{i, j})}{\sum_l \text{exp}(M_{i, l})}$, where $\quad M:=\frac{1}{\sqrt{D}}k(\mathbf{f}_{input})\cdot q(\mathbf{s}_{1:K}) ^T \in \mathbb{R}^{(H\times W)\times K}$, and $k$ and $q$ are learnable linear mappings $\mathbb{R}^{D \rightarrow D}$ as commonly used in the attention mechanism, and $\sqrt{D}$ is a fixed value for softmax temperature. With the calculated attention scores $\text{attn}_{i, j}$, image feature vectors $\mathbf{f}_{input}$ are aggregated via weighted mean: $\text{updates} : = \mathbf{W}^T\cdot v(\mathbf{f}_{input})\in \mathbb{R}^{K\times D}$, where $\mathbf{W}_{i, j}:= \text{attn}_{i,j}/(\sum_{l=1}^N\text{attn}_{l,j})$, and $v$ is also learnable linear mappings similar to $k$ and $q$. The update of slots in each iteration is completed via a learnable mapping parameterized by a Gated Recurrent Unit (GRU): $\mathbf{s}_{1:K} \leftarrow \text{GRU}(\mathbf{s}_{1:K}, \text{updates})$. The attention computation and updating are repeated 3 iterations to output final object-centric representations $\mathbf{s}_{1:K}$. Finally we obtain $K$ vectors $\mathbf{s}_k$ each of dimension 64. To infer probabilistic random variables from $\mathbf{s}_k$, a MLP is used to map $\mathbf{s}_k$ to $\mathbf{z}^s_k$. This MLP is implemented with two layers with the first layer followed by a RELU layer. To be emphasized, the MLP is shared across $\mathbf{s}_k$, to encourage common formats of objet representations. The obtained object-centric latent vector $\mathbf{z}^s_k$ is still with a dimension of 64. 

\textbf{Global Auto-Encoding Module.} To learn a global latent vector, the CNN backbone outputs $\mathbf{f}_x$ needs to be encoded by an encoder. Depending on the chosen prior distribution of the global latent vector, the encoder could have different structures. In the case that the global prior is Normal distribution, the encoder can be common ones used in vanilla VAE. Specifically, the $(H\times W) \times 64$ feature map is further flattened into one dimension, i.e., $(H\times W\times 64) \times 1$. Then a three-layer MLP, severing as an information bottleneck, reduces the dimension of obtained feature map to $\mathbf{z}^g$ of dimension $32 \times 1$. The obtained $\mathbf{z}^g$ can be decoded with deconvolutional neural nets back to the dimension of $(H\times W) \times 64$, trying to reconstruct the feature map. However, since the decoded feature map $\mathbf{f}$ is not used to recover image, rather generated object-centric latent vectors $\mathbf{z}^s_k$, there is no guarantee that $\mathbf{f}$ will be the same as $\mathbf{f}_x$. But with proper training, they should be close to each other. In summary, the auto-encoding structure is the same as commonly used VAE architecture. Another case for this global auto-encoding module is that a more powerful Strucdraw prior is used for the global latent vector learning. In that case, $\mathbf{z}^g$ is inferred autoregressively, the detail of such an encoder architecture could be found in \citep{jiang2020generative}. Along the path of global auto-encoding, the obtained $\mathbf{z}^g$ of dimension 32 is then fed into a slot attention module. This slot attention module has exactly the same architecture as the one on the first path. The two slot attention modules share parameters. 

\textbf{Object Component Decoder.} We choose the SBD decoder \citep{watters2019spatial} as part of the object component decoder in our model. Different from \citep{locatello2020object} and \citep{engelcke2019genesis} where a pure SBD is used, we combine SBD decoder with deconvolutional neural networks to balance the capacity of the decoder. Specifically, each object-centric latent vectors $\mathbf{z}^s_k$ of dimension 64 is first broadcast to a feature with shape $8\times 8\times 64$. Then this feature is decoded with deconvolutional neural nets with each layer having stride 2 and kernel size 5, to reconstruct an image-sized tensor with an additional channel as the mixing masks. The final output of the decoder has the shape $H\times W \times 4$. This decoder is shared across object-centric latent vectors $\mathbf{z}^s_k$.

\textbf{Hyperparameter for the KL term $D_{\text{KL}}\big[q_\phi(\mathbf{z}^{g} \mid \mathbf{x}) \mid\mid p_\theta(\mathbf{z}^{g})\big]$.} During training, we empirically find that  multiplying $D_{\text{KL}}\big[q_\phi(\mathbf{z}^{g} \mid \mathbf{x}) \mid\mid p_\theta(\mathbf{z}^{g})\big]$ with a small hyperparameter $\beta$ helps $\mathbf{z}^{g}$ to encode meaningful scene representations. When $\beta$ is too large, $\mathbf{z}^{g}$ tends to totally collapse to $p_\theta(\mathbf{z}^{g})$, i.e., normal distribution. In the experiments, for \emph{ObjectRoom}, $\beta$ is 0.01; for \emph{ShapeStacks}, $\beta$ is 0.1; and for \emph{Arrow Room}, $\beta$ is 0.1.

\textbf{Training Details.} Learning rate warm-up is important for object-centric representation learning as acknowledged by prior works. In the experiments, 10000 warm-up steps are used. For \emph{ObjectRoom}, the batch size is 64, and the learning rate is 0.0004; for \emph{ShapeStacks}, the batch size is 32, and the learning rate is 0.0001; and for \emph{Arrow Room}, the batch size is 32, and the learning rate is 0.0001 in the early training steps and is decreased to 0.00005 after object-centric representations show up for stable training purpose.

\end{document}